\documentclass[10pt,twocolumn,letterpaper]{article}
\pdfoutput=1

\usepackage{cvpr}              
\usepackage[accsupp]{axessibility}

\definecolor{cvprblue}{rgb}{0.21,0.49,0.74}
\usepackage[pagebackref,breaklinks,colorlinks,allcolors=cvprblue]{hyperref}
\usepackage{siunitx}
\usepackage{multirow}
\usepackage{pifont}
\usepackage[most]{tcolorbox}
\newtcblisting{promptbox}{
  listing only, breakable, arc=2mm, boxrule=0.4pt,
  colback=gray!3, colframe=gray!40,
  listing options={basicstyle=\ttfamily\small, breaklines=true}
}
\usepackage{algorithm}
\usepackage{algpseudocode}

\algrenewcommand\algorithmicrequire{\textbf{Input:}}
\algrenewcommand\algorithmicensure{\textbf{Output:}}

\def\paperID{40844} 
\def\confName{CVPR}
\def\confYear{2026}

\title{rPPG-VQA: A Video Quality Assessment Framework for Unsupervised rPPG Training}

\author{Tianyang Dai, Ming Chang, Yan Chen, Yang Hu\thanks{Corresponding author.}\\
University of Science and Technology of China\\
{\tt\small \{daitianyang, mingchang\}@mail.ustc.edu.cn, \{eecyan, eeyhu\}@ustc.edu.cn}
}

\begin{document}
\maketitle
\begin{abstract}
Unsupervised remote photoplethysmography (rPPG) promises to leverage unlabeled video data, but its potential is hindered by a critical challenge: training on low-quality ``in-the-wild'' videos severely degrades model performance. An essential step missing here is to assess the suitability of the videos for rPPG model learning before using them for the task. Existing video quality assessment (VQA) methods are mainly designed for human perception and not directly applicable to the above purpose. In this work, we propose rPPG-VQA, a novel framework for assessing video suitability for rPPG. We integrate signal-level and scene-level analyses and design a dual-branch assessment architecture. The signal-level branch evaluates the physiological signal quality of the videos via robust signal-to-noise ratio (SNR) estimation with a multi-method consensus mechanism, and the scene-level branch uses a multimodal large language model (MLLM) to identify interferences like motion and unstable lighting. Furthermore, we propose a two-stage adaptive sampling (TAS) strategy that utilizes the quality score to curate optimal training datasets. Experiments show that by training on large-scale, ``in-the-wild'' videos filtered by our framework, we can develop unsupervised rPPG models that achieve a substantial improvement in accuracy on standard benchmarks. Our code is available at \url{https://github.com/Tianyang-Dai/rPPG-VQA}.
\end{abstract}    
\section{Introduction}
\label{sec:Introduction}


\begin{figure}[t]
    \centering
    \includegraphics[width=0.9\linewidth]{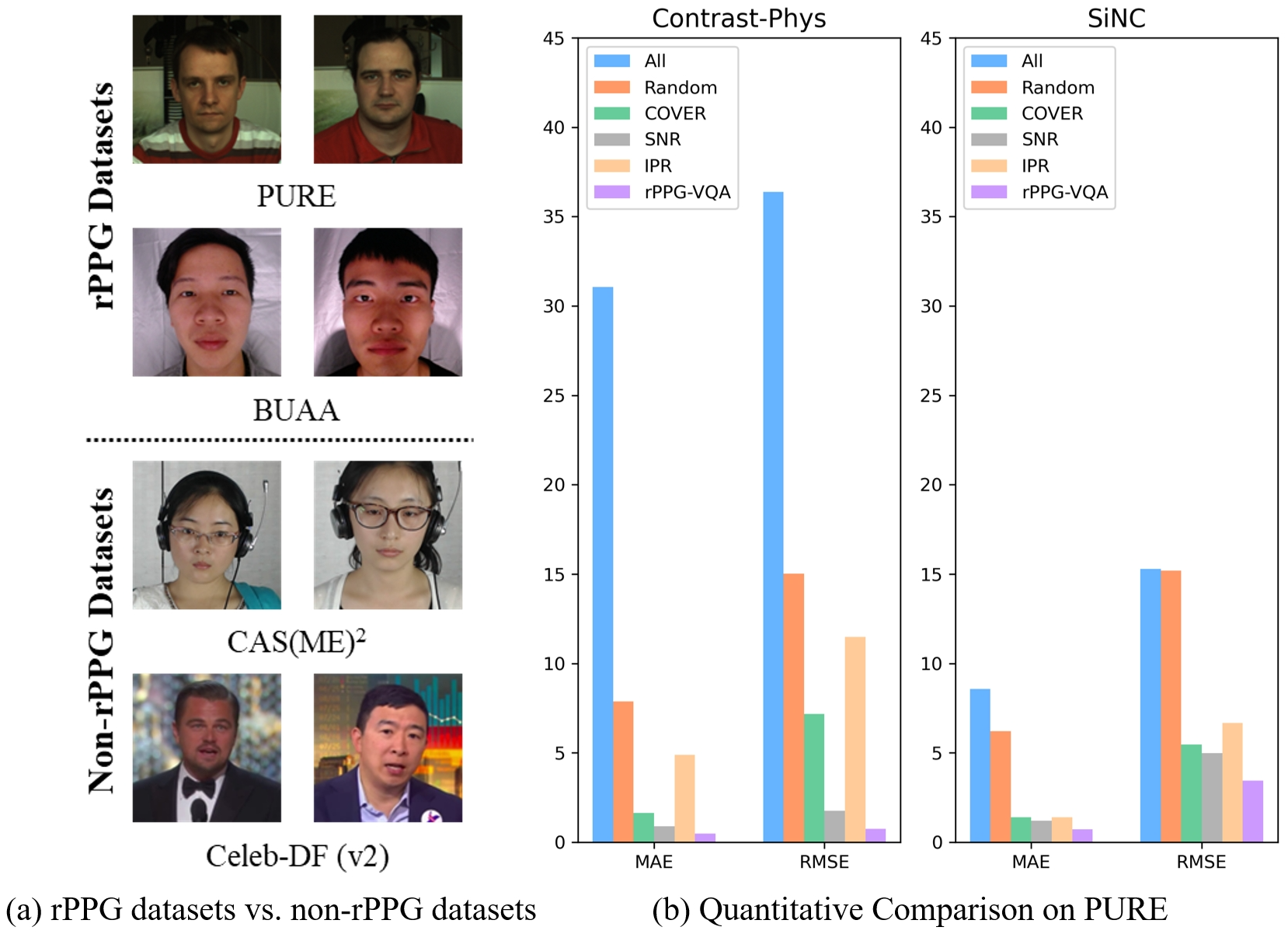}
    \caption{\textbf{Dataset categorization and performance comparison for unsupervised rPPG.} (a) Classification of datasets into labeled rPPG benchmarks and unlabeled ``in-the-wild'' corpora. (b) HR estimation errors on the PURE test set using different sampling criteria. ``All'' and ``Random'' denote the full and randomly sampled combined dataset (CAS(ME)\textsuperscript{2} and Celeb-DF (v2)), respectively.}
    \label{fig:illustration}
\end{figure}

Remote photoplethysmography (rPPG) enables contactless heart rate (HR) measurement using standard cameras, circumventing the need for skin-contact sensors like electrocardiograms (ECG) and photoplethysmography (PPG) in applications ranging from neonatal care to remote healthcare~\cite{chen2018video, mcduff2023camera, shi2019atrial}. While deep learning methods have proven highly effective at extracting the weak physiological signal from noisy video data, their reliance on supervised learning presents a significant bottleneck~\cite{liu2023rppg, niu2019rhythmnet}. This dependency requires large-scale datasets with synchronized video and ground-truth physiological signals, resources that are expensive and labor-intensive to acquire~\cite{yu2019remote}. Meanwhile, the generalization ability of supervised models to novel, unseen scenarios is severely compromised~\cite{stricker2014non, niu2018vipl, bobbia2019unsupervised}, spurring the development of unsupervised rPPG methods as a crucial research frontier.

Unsupervised rPPG reduces reliance on labeled training data~\cite{gideon2021way, sun2022contrast, speth2023non, yue2023facial, zhang2024self}. However, their development has predominantly focused on methodological innovation, often overlooking a crucial data-centric challenge: not all videos are suitable for rPPG analysis. Current research typically validates new methods using controlled, high-quality datasets, yet the true value of unsupervised learning lies in its ability to harness unconstrained data. In such real-world settings, severe motion and illumination noise can overwhelm the subtle physiological signal and corrupt the training process~\cite{speth2023non}. Furthermore, a valid rPPG signal may be intrinsically weak or entirely absent due to factors like aggressive video compression or the prevalence of AI-generated facial videos, which lack a genuine physiological basis. Training on such unfiltered data can degrade model performance (as shown in \Cref{fig:illustration}), highlighting a significant gap in the literature: the absence of a robust video quality assessment framework tailored to the specific needs of rPPG.

The rapid growth of various video data has made video quality assessment (VQA) an important task in video processing and analysis. However, traditional VQA was developed for applications like video streaming and compression, with the primary goal of quantifying perceptual quality for the human vision system (HVS)~\cite{li2024misc, li2024g, li2024q}. These methods evaluate visual distortions such as blur and compression artifacts using metrics like PSNR and SSIM, which are not designed to assess the integrity of underlying physiological signals~\cite{zheng2024video, li2025image, wang2003multiscale}. This creates a fundamental disconnect, as the criteria for the HVS and a machine vision system (MVS) performing rPPG are different. A video that appears visually clear to a human may be useless for rPPG if the physiological signal is destroyed or missing, while a visually degraded video might still contain an extractable physiological signal.

Even rPPG-specific quality metrics alone, such as signal-to-noise ratio (SNR), are unreliable for this task. Such metrics are easily deceived by non-physiological phenomena, for instance, periodic camera flashes that mimic a cardiac signal, which can artificially inflate quality scores. This fundamental flaw stems from their inability to verify a signal's physiological origin, as they lack the scene context to distinguish true biometric data from confounding artifacts. A robust VQA mechanism for rPPG must therefore move beyond simple signal-centric metrics.

To address this challenge, we propose rPPG-VQA, a novel dual-branch framework that assesses video quality by concurrently analyzing signal-level integrity and identifying scene-level interferences.
The signal-level noise perception branch tackles the unreliability of standalone SNR scores. Based on the principle that a true physiological signal should be method-agnostic, this branch computes a consensus score from the SNR estimates of multiple, diverse rPPG algorithms, mitigating the biases of any single method.
The scene-level noise perception branch identifies interferences that elude traditional metrics. It leverages a multimodal large language model (MLLM) to perform human-like scene reasoning, detecting complex interferences such as unstable lighting, erratic motion, and prohibitive camera artifacts.
Meanwhile, obtaining a reliable quality score for each video is only the first step. The subsequent challenge lies in leveraging the score to curate an optimal training dataset from a vast, unvetted video pool. For this purpose, we further introduce a two-stage adaptive sampling (TAS) strategy, which first filters out low-quality videos and then applies duration-aware probabilistic sampling to construct a training dataset that optimally balances quality, diversity, and efficiency. \Cref{fig:illustration} provides an overview of the dataset categorization challenge and the quantitative improvements achieved by our proposed framework.

In summary, the contributions of this paper are as follows:
\begin{itemize}
    \item We address the problem of video quality assessment for unsupervised rPPG. To our knowledge, this is the first systematic study of this pivotal problem for the task. Unlike most existing VQA works which mainly focus on perceptual quality, our primary concern is the physiological signal integrity in the videos.
    \item We propose rPPG-VQA, a novel framework that uniquely combines signal-level noise perception with scene-level interference analysis to assess video suitability for rPPG.
    \item We introduce the two-stage adaptive sampling strategy, a practical method that uses our quality score to curate effective training datasets from unvetted, in-the-wild sources.
    \item We demonstrate the efficacy of our framework by training unsupervised rPPG models on large-scale, ``in-the-wild'' video data which demonstrate significant accuracy enhancement.
\end{itemize}

\section{Related work}
\label{sec:Related work}

\subsection{General-Purpose VQA}

Conventional VQA is an established field dedicated to modeling human perception of video quality. These frameworks are designed to quantify the impact of distortions, such as compression artifacts or frame-rate fluctuations, on a human observer.

Methodologically, these methods are categorized by their reliance on a reference video. Full-reference (FR) methods, like PSNR, SSIM~\cite{wang2004image}, and VMAF~\cite{rassool2017vmaf}, evaluate quality by comparing a distorted video against its pristine original at the pixel, structural, or textural levels. In contrast, no-reference (NR) or blind methods such as NIQE~\cite{mittal2012no} and BRISQUE~\cite{mittal2012making} predict perceptual quality from intrinsic natural scene statistics, requiring no reference video. Recent innovations in blind VQA have produced modular designs that better model human sensitivity to specific factors like spatial resolution and temporal changes~\cite{wen2024modular}.

However, this inherent focus on the HVS renders such general-purpose models unsuitable for specialized machine vision tasks like rPPG. The core objective of conventional VQA is to assess perceptual clarity for a human viewer, prioritizing aspects of overall visual fidelity~\cite{ou2014q, janowski2010qoe, ou2010modeling, zheng2022no, yi2021attention, wu2022fast, wu2023exploring}. In contrast, a VQA for rPPG must determine if a stable physiological signal is extractable. This depends on an entirely different set of criteria, beginning with the intrinsic viability of a physiological signal and extending to factors like illumination stability, motion smoothness, and the presence of unobscured skin regions. They all are factors that standard VQA frameworks are not designed to measure.

\subsection{Signal Quality Assessment for rPPG}

Current methods to assessing rPPG data quality are largely ad-hoc and operate only at the signal level, making them unsuitable for pre-emptive data filtering in unsupervised learning pipelines. These methods typically infer quality post-hoc by analyzing characteristics of an already extracted signal, such as its temporal morphology~\cite{fischer2016algorithm, fischer2017extended}, frequency-domain energy distribution~\cite{wang2016quality}, or SNR~\cite{speth2023non, elgendi2016optimal, wu2025semi}. Other techniques, including those based on image-region variations~\cite{fallet2017imaging} or uncertainty quantification~\cite{song2023uncertainty}, still presuppose a processed signal. Because they cannot evaluate a raw video's intrinsic suitability for rPPG training, they are inadequate for curating large, unlabeled datasets.

Beyond their post-hoc nature, existing metrics overlook key factors that influence signal extraction, such as the interplay between skin tone and illumination or the effects of camera sensor noise. Notably, they often lack a direct correlation with the ultimate performance metric: heart rate estimation accuracy. This disconnect is a significant shortcoming, as variations in camera type, lighting, and motion create substantial domain shifts that degrade model generalization, particularly when training data lacks diversity~\cite{zhang2025advancing}.

The absence of a task-specific, pre-emptive quality metric has created a major bottleneck for the field, hindering both reliable signal extraction and the cross-domain generalization of rPPG models. Consequently, even state-of-the-art unsupervised methods remain tethered to specialized, curated datasets~\cite{speth2023non}, unable to leverage the vast amount of unlabeled ``in-the-wild'' facial video. Developing a reference-free VQA framework tailored specifically for rPPG is therefore essential to unlock the potential of large-scale unsupervised learning.

\section{Method}
\label{sec:Method}

\begin{figure*}[t]
    \centering
    \includegraphics[width=0.8\linewidth]{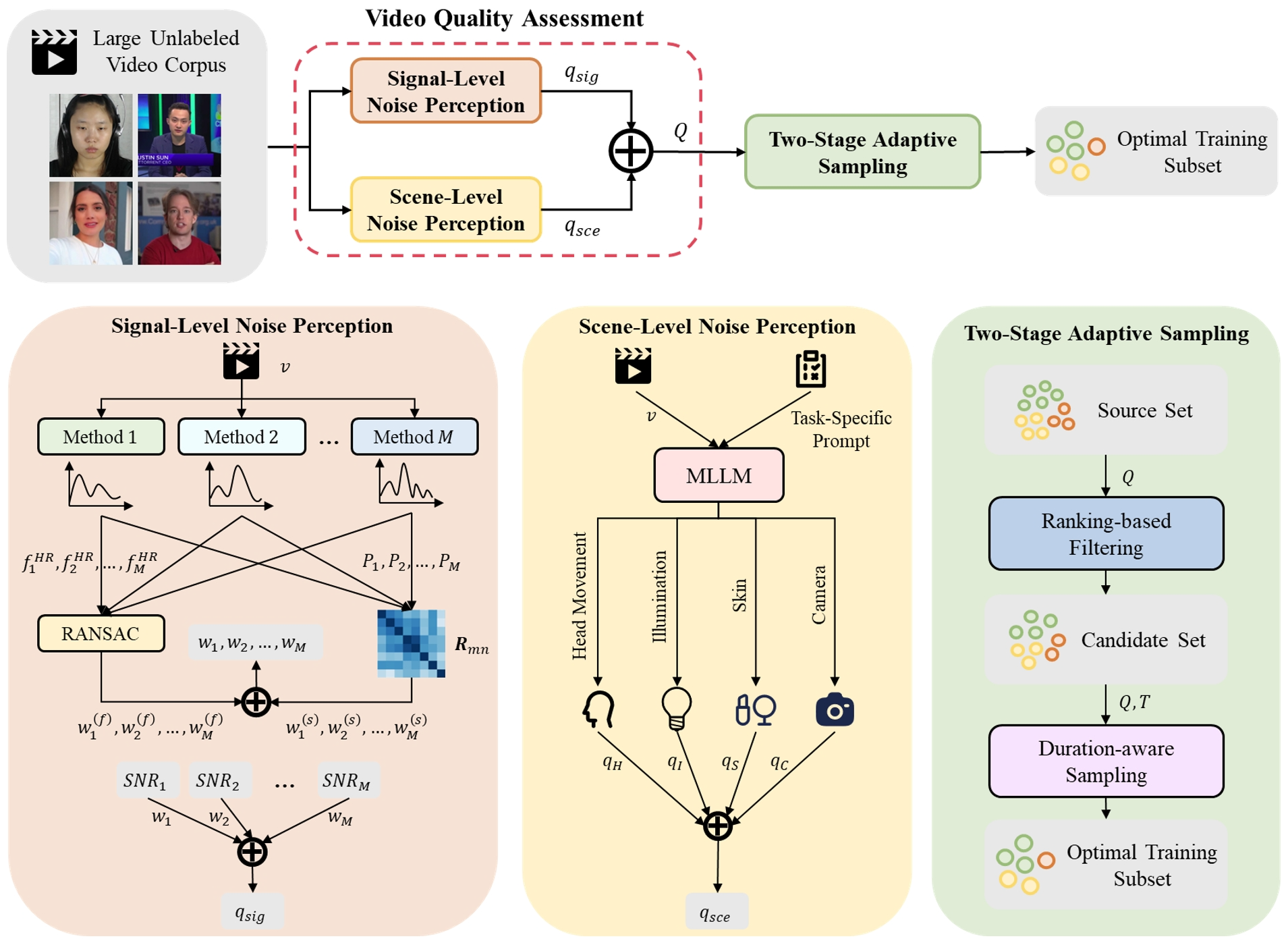}
    \caption{\textbf{An overview of our proposed rPPG-VQA framework.} The method employs a dual-branch architecture to assess video quality. The signal-level noise perception branch evaluates physiological signal integrity by fusing SNR estimates from multiple rPPG algorithms based on their consensus. The scene-level noise perception branch uses an MLLM to score interferences like motion and illumination. The two assessments are combined into a unified quality score, which guides a two-stage adaptive sampling strategy to construct the final target training set.}
    \label{fig:framework}
\end{figure*}

\subsection{Problem Formulation}

Our primary goal is to assign a quality score $Q\left( {{v_i}} \right)$ to each video ${v_i}$ in a large, unlabeled source corpus $\mathcal{D}_{src}$. This score must quantify two key aspects of the video: (i) the integrity of the underlying physiological signal and (ii) the severity of scene-level interferences. The resulting scores are then used to guide the construction of a target training set $\mathcal{D}_{tgt}$.

To achieve this, we introduce rPPG-VQA, a novel dual-branch framework as illustrated in \Cref{fig:framework}. The first branch, signal-level noise perception, quantifies signal integrity by assessing the consensus among SNR estimates from multiple, diverse rPPG algorithms. The second branch, scene-level noise perception, leverages an MLLM to identify scene interferences, such as motion, illumination, and skin optics, that are beyond the reach of traditional metrics. These two assessments are fused into a holistic quality score, which directs a two-stage adaptive sampling strategy to construct the final dataset $\mathcal{D}_{tgt}$.

\subsection{Signal-Level Noise Perception Branch}
\label{subsec:Signal-Level Noise Perception Branch}

Assessing the intrinsic quality of an rPPG signal typically involves extracting the signal first, then evaluating its properties~\cite{speth2023non, wu2025semi}. Although learning-based extraction methods demonstrate strong performance on specific benchmarks, they often struggle with generalization to datasets with different characteristics. Conversely, conventional signal processing-based rPPG methods (\eg, GREEN~\cite{verkruysse2008remote}, ICA~\cite{poh2010advancements}, CHROM~\cite{de2013robust}, LGI~\cite{pilz2018local}, PBV~\cite{de2014improved}, POS~\cite{wang2016algorithmic}, and OMIT~\cite{casado2023face2ppg}) require no training data but are often fragile in real-world conditions with motion and illumination artifacts. To achieve a more robust quality estimate and overcome the limitations of any single method, a logical step is to fuse the outputs of multiple, diverse algorithms. However, simple averaging is unreliable, as one poorly performing method can corrupt the aggregate result. We therefore propose a fusion strategy centered on multi-method consistency, which rewards agreement among methods and systematically penalizes outliers.

To quantify the quality of a signal from a given method, we first compute its SNR. The SNR is computed from the power spectral density (PSD) $P\left( f \right)$ of the rPPG signal extracted from a specific video $v_i$, as proposed in~\cite{de2013robust}. The metric is the decibel ratio of the power within a signal band versus a noise band over the frequency range $\left[ {\SI{0.75}{\hertz},\SI{2.5}{\hertz}} \right]$. The signal band ${B_{sig}}$ captures energy in a narrow window ($\Delta  = \SI{0.1}{\hertz}$) around the fundamental (${f_1}$) and second harmonic (${f_2} = 2{f_1}$) frequencies of the estimated heart rate:
\begin{equation}
    B_{sig} \;=\; [f_1-\Delta,\, f_1+\Delta] \,\cup\, [f_2-\Delta,\, f_2+\Delta]
\end{equation}
The noise band ${B_{noise}}$ comprises the remainder of the analysis range. The final SNR is given by:
\begin{equation}
    SNR = 10{\log _{10}}\left( {\frac{{\sum\limits_{f \in {B_{sig}}} {P\left( f \right)} }}{{\sum\limits_{f \in {B_{noise}}} {P\left( f \right)} }}} \right)
\end{equation}

With the general SNR calculation defined, let $SN{R_{i,m}}$ denote the SNR computed for video ${v_i}$ using the $m$-th extraction method. We formalize our consistency-based fusion as a weighted sum of these SNR values from $M$ diverse methods. The signal-level quality score for a video ${v_i}$ is defined as:
\begin{equation}
    {q_{sig}}\left( {{v_i}} \right) = \sum\limits_{m = 1}^M {{w_{i,m}}SN{R_{i,m}}}
\end{equation}
where the data-driven fusion weight ${w_{i,m}}$ reflects the agreement of method $m$ with the consensus for video $v_i$. To capture different facets of agreement, we decompose ${w_{i,m}}$ into two complementary, normalized terms:
\begin{equation}
    {w_{i,m}} = w_{i,m}^{\left( f \right)} + w_{i,m}^{\left( s \right)}
\end{equation}
where $w_{i,m}^{\left( f \right)}$ and $w_{i,m}^{\left( s \right)}$ are the frequency consistency weight and the spectral correlation weight, respectively.

\subsubsection{Frequency Consistency Weight}

The frequency consistency weight $w_{i,m}^{\left( f \right)}$ rewards methods that agree on the estimated heart rate for video $v_i$. The premise is that accurate methods should converge on the same physiological frequency. First, for each video $v_i$, we identify each method's estimated heart rate frequency $f_{i,m}^{HR}$ by finding the peak in its PSD ${P_{i,m}}\left( f \right)$:
\begin{equation}
    f_{i,m}^{HR} = \mathop {\arg \max }\limits_{f \in \left[ {{f_{\min }},{f_{\max }}} \right]} {P_{i,m}}\left( f \right)
\end{equation}

To establish a robust reference frequency for video $v_i$, we use the random sample consensus (RANSAC)~\cite{fischler1981random} algorithm on the set of individual estimates. This prevents outliers from skewing a simple average:
\begin{equation}
    f_{i,cons}^{HR} = RANSAC\left( {\left\{ {f_{i,1}^{HR},f_{i,2}^{HR},...,f_{i,M}^{HR}} \right\},\epsilon } \right)
\end{equation}
where $\epsilon$ is the inlier tolerance threshold.

Finally, we map the deviation of each method's estimate from the consensus to a weight using a Gaussian kernel. This function smoothly penalizes methods as their estimates diverge:
\begin{equation}
    w_{i,m}^{\left( f \right)} = \exp \left( { - \frac{{{{\left( {f_{i,m}^{HR} - f_{i,cons}^{HR}} \right)}^2}}}{{2\sigma _f^2}}} \right)
\end{equation}
where $\sigma_f$ controls the sharpness of the penalty. Consequently, methods aligned with the consensus receive higher weights.

\subsubsection{Spectral Correlation Weight}

Beyond a single peak frequency, the overall shape of the PSD also contains valuable information. The spectral correlation weight $w_{i,m}^{\left( s \right)}$, rewards methods whose PSDs are morphologically similar to their peers, based on the assumption that reliable methods should share a common spectral signature.

For each video $v_i$, we first compute a spectral correlation matrix ${{\bf{R}}_i} \in {\mathbb{R}^{M \times M}}$, where each element ${{\bf{R}}_{i,mn}}$ is the Pearson correlation between the normalized PSDs of methods $m$ and $n$:
\begin{equation}
    {{\bf{R}}_{i,mn}} = \rho \left( {{P_{i,m}},{P_{i,n}}} \right)
\end{equation}

The weight for method $m$ is then derived by averaging its squared correlations with all other methods, which rewards broad spectral consensus:
\begin{equation}
    w_{i,m}^{\left( s \right)} = \frac{1}{{M - 1}}\sum\limits_{n \ne m} {{\bf{R}}_{i,mn}^2}
\end{equation}

This method systematically down-weights methods whose extracted signals are with idiosyncratic or noisy spectral shapes, leading to a more robust fusion.

\subsection{Scene-Level Noise Perception Branch}
\label{subsec:Scene-Level Noise Perception Branch}

The quality of rPPG signals is often degraded by artifacts such as head movements, illumination variations, and diverse skin characteristics. Conventional assessment methods rely on low-level, handcrafted features~\cite{zhang2025advancing} that lack the capacity to interpret these scene information. For instance, they struggle to differentiate benign facial expressions from disruptive head movements or to assess the perceptual impact of non-uniform lighting.

To overcome these limitations, our scene-level noise perception branch leverages the emerging power of recent MLLMs. This method directly evaluates video quality by assessing four primary noise dimensions using task-specific prompts, thereby eliminating the need for handcrafted features or architectural modifications. These dimensions are summarized in \Cref{tab:noise_scores}.

\begin{table}
    \caption{\textbf{Scene noise dimensions and scoring scales for MLLM-based assessment.} Higher scores denote better quality.}
    \label{tab:noise_scores}
    \centering
    \scalebox{0.9}{
    \begin{tabular}{cc}
        \toprule
        Dimension            & Score       \\
        \midrule
        Head Movement Noise  & $\left\{ {0,1,2,3} \right\}$ \\
        Illumination Noise   & $\left\{ {0,1,2,3} \right\}$ \\
        Skin Noise   & $\left\{ {0,1,2} \right\}$   \\
        Camera Noise & ${\gamma _i} \cdot \left\{ {0,1,2} \right\}$   \\ 
        \bottomrule
    \end{tabular}
    }
\end{table}

For a given video $v_i$ and a noise-specific prompt ${I_j}$, the employed MLLM generates a quality score ${q_j}\left( {{v_i}} \right) = MLLM\left( {{v_i},{I_j}} \right)$, where $j \in \left\{ {H,I,S,C} \right\}$ corresponds to:
\begin{itemize}
    \item \textbf{Head Movement (H):} Sceneally evaluates motion severity, distinguishing disruptive movements from natural expressions.
    \item \textbf{Illumination (I):} Assesses the perceptual impact of lighting instability (\eg, fluctuations, over/underexposure).
    \item \textbf{Skin (S):} Accounts for optical interferences such as skin tone, facial hair, or makeup that affect signal reflection.
    \item \textbf{Camera (C):} Evaluates artifacts from sensor noise and compression. This score is uniquely scaled by a frame rate compensation factor ${\gamma _i} = \frac{{\min \left( {f_i^{orig},30} \right)}}{{30}}$, where $f^{orig}_i$ is the video's original frame rate. This factor penalizes information distortion from upsampling low-frame-rate videos. The final score is ${q_C}\left( {{v_i}} \right) = {\gamma _i} \cdot MLLM\left( {{v_i},{I_C}} \right)$.
\end{itemize}

The final scene-level quality score ${q_{sce}}$ is a sum of the individual scores:
\begin{equation}
    {q_{sce}}\left( {{v_i}} \right) = \sum\limits_{j \in \left\{ {H,I,S,C} \right\}} {{q_j}\left( {{v_i}} \right)}
\end{equation}

\subsection{Unified Video Quality Score}
\label{subsec:Unified Video Quality Score}

To compute the unified video quality score $Q\left( {{v_i}} \right)$, we first truncate outliers using the interquartile range (IQR) method to ensure robust scaling. We then normalize the signal-level score ${q_{sig}}\left( {{v_i}} \right)$ and the scene-level score ${q_{sce}}\left( {{v_i}} \right)$ to the range $\left[ {0,1} \right]$ via min-max scaling, and combine them as a weighted average:
\begin{equation}
    Q\left( {{v_i}} \right) = \alpha {q_{sig}}\left( {{v_i}} \right) + \left( {1 - \alpha } \right){q_{sce}}\left( {{v_i}} \right)
\end{equation}
where $\alpha  \in \left[ {0,1} \right]$ balances the importance of physiological signal integrity against the severity of scene interferences.

\subsection{Data Sampling}
\label{subsec:Data Sampling}

Having assigned a quality score to each video, the next step is to devise a sampling strategy to construct the target training set $\mathcal{D}_{tgt}$. Common strategies like Top/Bottom $K$ selection are crude, collapsing rich scores into a hard cutoff. Another method is weighted random sampling (WRS)~\cite{efraimidis2006weighted}, which makes each video's sampling probability proportional to its quality score, tends to oversample short clips by ignoring their duration.

To address these limitations, we propose a two-stage adaptive sampling (TAS) strategy. TAS is adaptive because it dynamically adjusts sampling probabilities based on both video quality and duration. It first employs ranking-based filtering to create a high-quality candidate pool, then performs duration-aware probabilistic sampling to balance data quality, diversity, and data efficiency.

\subsubsection{Ranking-Based Filtering}

First, all videos are sorted in descending order of their quality scores $Q\left( {{v_i}} \right)$. The top-ranked videos are selected to form a candidate set ${\mathcal{D}_{cand}}$, whose size is a multiple of the target training set size:
\begin{equation}
    \left| {{\mathcal{D}_{cand}}} \right| = \eta \left| {{\mathcal{D}_{tgt}}} \right|, \quad \eta > 1
\end{equation}

\subsubsection{Duration-Aware Sampling}

Next, we perform probabilistic sampling from ${{\mathcal{D}_{cand}}}$. The sampling probability $p\left( {{v_i}} \right)$ for each video incorporates both its quality score $Q\left( {{v_i}} \right)$ and duration ${T_i}$ via a softmax distribution:
\begin{equation}
    p\left( {{v_i}} \right) = \frac{{\exp \left( {{\lambda _i}Q\left( {{v_i}} \right)/\tau } \right)}}{{\sum\limits_{j \in {\mathcal{D}_{cand}}} {\exp \left( {{\lambda _j}Q\left( {{v_j}} \right)/\tau } \right)} }}
\end{equation}
where $\tau$ is a temperature parameter and $\lambda_i$ is a duration adjustment factor:
\begin{equation}
    {\lambda _i} = \log \left( {1 + \frac{{{T_i}}}{{\bar T}}} \right)
\end{equation}
with $\bar{T}$ being the average video duration in ${\mathcal{D}_{cand}}$. This logarithmic formulation ensures the influence of duration grows sublinearly, preventing exceptionally long videos from dominating the sampling process.

The expected sampling count $r\left( {{v_i}} \right)$ for each video is then:
\begin{equation}
    r\left( {{v_i}} \right) = p\left( {{v_i}} \right)\left| {{D_{tgt}}} \right|
\end{equation}

Since $r\left( {{v_i}} \right)$ is a real number, we use stochastic rounding to determine the final integer count~\cite{xi2025samplemix}:
\begin{itemize}
    \item \textbf{Integer Part:} Each video $v_i$ is sampled $\left\lfloor {r\left( {{v_i}} \right)} \right\rfloor$ times.
    \item \textbf{Fractional Part:} An additional sample of $v_i$ is taken with probability $r\left( {{v_i}} \right) - \left\lfloor {r\left( {{v_i}} \right)} \right\rfloor$.
\end{itemize}

Finally, a random clip is extracted each time a video is selected, and this process is repeated until the target training set ${\mathcal{D}_{tgt}}$ is constructed.

\section{Experiments}
\label{sec:Experiments}

\subsection{Experimental Setup}

\noindent \textbf{Training Sets.} We construct a large-scale, unlabeled ``in-the-wild'' video corpus from two public datasets. CAS(ME)\textsuperscript{2}~\cite{qu2017cas} contains videos of 22 subjects exhibiting diverse facial dynamics, from subtle micro-expressions to significant head movements, making it a challenging benchmark for motion robustness. Celeb-DF (v2)~\cite{li2020celeb} is a large-scale deepfake detection dataset. We use its 890 authentic videos, which feature diverse subjects in varied lighting and motion conditions. Neither dataset is originally collected for the rPPG task and therefore they contain no ground-truth pyhsiological signal labels.

\noindent \textbf{Test Sets.} We evaluate model performance on two standard rPPG benchmarks. The PURE dataset~\cite{stricker2014non} contains 60 videos of 10 subjects performing six motion tasks (\eg, stationary, talking), ideal for assessing motion robustness. The BUAA-MIHR dataset~\cite{xi2020image} contains 142 videos from 13 subjects under varying illumination levels. Following common practice, videos recorded below 10 lux are excluded.

\noindent \textbf{Unsupervised rPPG Methods.} To demonstrate the general applicability of our framework, we evaluate it on three unsupervised rPPG methods: Gideon21~\cite{gideon2021way}, Contrast-Phys~\cite{sun2022contrast}, and SiNC~\cite{speth2023non}.

\noindent \textbf{Comparison Baselines.} We benchmark our method against baselines for both quality assessment and data sampling. For quality criteria, we compare our rPPG-VQA against: 1) COVER~\cite{he2024cover}, a state-of-the-art general video quality metric; and 2) two rPPG-specific metrics, SNR and IPR, calculated from signals predicted by the POS algorithm~\cite{wang2016algorithmic}, a representative and top-performing conventional method used for this purpose in prior works~\cite{speth2023non, wu2025semi}. For sampling strategies, we compare our TAS against random selection, Top/Bottom $K$ selection, and WRS.

\noindent \textbf{Evaluation Metrics.} Following standard rPPG protocols~\cite{gideon2021way, sun2022contrast, speth2023non}, we assess HR estimation performance using mean absolute error (MAE), root mean square error (RMSE), and the Pearson correlation coefficient (R).

\noindent \textbf{Implementation Details.} Our framework is implemented in PyTorch and run on an NVIDIA V100 GPU. All videos and signals are resampled to 30 fps. The signal-level noise perception branch integrates seven conventional rPPG methods: GREEN~\cite{verkruysse2008remote}, ICA~\cite{poh2010advancements}, CHROM~\cite{de2013robust}, LGI~\cite{pilz2018local}, PBV~\cite{de2014improved}, POS~\cite{wang2016algorithmic}, and OMIT~\cite{casado2023face2ppg}. These methods, mostly computationally inexpensive and parallelizable, introduce minimal overhead for our offline analysis. The RANSAC inlier threshold is set to $\epsilon = 5 \ \text{bpm}$, and the Gaussian kernel's standard deviation ${\sigma _f}$ is dynamically set to the standard deviation of the HR estimates from the conventional methods. The scene-level noise perception branch employs Qwen3-VL~\cite{bai2025qwen3}, with prompts for the four noise dimensions detailed in the supplementary material.

For the training configuration, the final quality score $Q$ is computed with a fusion weight of $\alpha  = 0.8$. Our TAS strategy uses a candidate set ratio of $\eta  = 2.0$ and a softmax temperature of $\tau  = 1.0$. The source data pool is formed from the CAS(ME)\textsuperscript{2} and Celeb-DF (v2) datasets. We construct a target training set ${\mathcal{D}_{tgt}}$ of 140 videos by including all videos from CAS(ME)\textsuperscript{2} and selecting a subset from Celeb-DF (v2) using our proposed method. This balances the high signal quality of CAS(ME)\textsuperscript{2} with the scenic diversity of Celeb-DF (v2). All unsupervised rPPG models are trained on ${\mathcal{D}_{tgt}}$ with their original hyperparameters and subsequently evaluated on the PURE and BUAA test sets.

\begin{table*}
\caption{\textbf{Comparison of quality criteria and sampling strategies for training data selection, evaluated across various unsupervised rPPG methods on the PURE dataset.} The best performing combination for each metric is highlighted in bold, while the second-best is underlined.}
\label{tab:results_pure}
\centering
\scalebox{0.8}{
\begin{tabular}{cccccccccccc}
\toprule
\multirow{2}{*}{\textbf{Training Set}} & \multirow{2}{*}{\textbf{Sampling}} & \multirow{2}{*}{\textbf{Quality Criterion}} & \multicolumn{3}{c}{\textbf{Gideon21}} & \multicolumn{3}{c}{\textbf{Contrast-Phys}} & \multicolumn{3}{c}{\textbf{SiNC}} \\ \cline{4-12} 
 &  &  & MAE$\downarrow$ & RMSE$\downarrow$ & R$\uparrow$ & MAE$\downarrow$ & RMSE$\downarrow$ & R$\uparrow$ & MAE$\downarrow$ & RMSE$\downarrow$ & R$\uparrow$ \\ \midrule
CAS(ME)\textsuperscript{2} & - & - & 1.80 & 2.24 & \underline{0.99} & 1.72 & 2.74 & 0.98 & 2.00 & 8.16 & 0.93 \\ \midrule
Celeb-DF (v2) & - & - & 11.94 & 13.92 & 0.36 & 37.34 & 40.82 & -0.14 & 23.12 & 28.45 & 0.03 \\ \midrule
\multirow{11}{*}{\begin{tabular}[c]{@{}c@{}}CAS(ME)\textsuperscript{2}\\ + Celeb-DF (v2)\end{tabular}} & - & - & 6.91 & 12.26 & 0.57 & 31.07 & 36.37 & -0.30 & 8.58 & 15.28 & 0.39 \\ \cline{2-12} 
 & Random & - & 4.04 & 12.02 & 0.64 & 7.86 & 15.02 & 0.39 & 6.20 & 15.20 & 0.56 \\ \cline{2-12} 
 & \multirow{4}{*}{Top/Bottom $K$} & COVER & 0.46 & 0.60 & \textbf{1.00} & 2.56 & 10.07 & 0.72 & 1.69 & 6.76 & 0.88 \\
 &  & SNR & 0.47 & 0.58 & \textbf{1.00} & 2.34 & 8.32 & 0.82 & 0.76 & 4.23 & 0.95 \\
 &  & IPR & 0.57 & 0.75 & \textbf{1.00} & 4.63 & 10.53 & 0.71 & 1.52 & 6.54 & 0.89 \\
 &  & \textbf{rPPG-VQA} & 0.42 & 0.50 & \textbf{1.00} & 1.36 & 4.05 & 0.96 & 0.87 & 4.33 & 0.95 \\ \cline{2-12} 
 & \multirow{4}{*}{WRS} & COVER & 0.50 & 0.58 & \textbf{1.00} & 1.64 & 7.17 & 0.86 & 1.39 & 5.46 & 0.92 \\
 &  & SNR & 0.42 & 0.55 & \textbf{1.00} & 0.88 & 1.76 & \underline{0.99} & 1.19 & 4.97 & 0.94 \\
 &  & IPR & 0.57 & 0.72 & \textbf{1.00} & 4.89 & 11.49 & 0.67 & 1.39 & 6.67 & 0.88 \\
 &  & \textbf{rPPG-VQA} & \underline{0.37} & \underline{0.47} & \textbf{1.00} & \underline{0.71} & \underline{1.60} & \underline{0.99} & \underline{0.74} & \underline{4.12} & \underline{0.96} \\ \cline{2-12} 
 & TAS & \textbf{rPPG-VQA} & \textbf{0.36} & \textbf{0.44} & \textbf{1.00} & \textbf{0.47} & \textbf{0.74} & \textbf{1.00} & \textbf{0.72} & \textbf{3.43} & \textbf{0.97} \\ \bottomrule
\end{tabular}
}
\end{table*}

\begin{table*}
\caption{\textbf{Comparison of quality criteria and sampling strategies for training data selection, evaluated across various unsupervised rPPG methods on the BUAA dataset.}}
\label{tab:results_buaa}
\centering
\scalebox{0.8}{
\begin{tabular}{cccccccccccc}
\toprule
\multirow{2}{*}{\textbf{Training Set}} & \multirow{2}{*}{\textbf{Sampling}} & \multirow{2}{*}{\textbf{Quality Criterion}} & \multicolumn{3}{c}{\textbf{Gideon21}} & \multicolumn{3}{c}{\textbf{Contrast-Phys}} & \multicolumn{3}{c}{\textbf{SiNC}} \\ \cline{4-12} 
 &  &  & MAE$\downarrow$ & RMSE$\downarrow$ & R$\uparrow$ & MAE$\downarrow$ & RMSE$\downarrow$ & R$\uparrow$ & MAE$\downarrow$ & RMSE$\downarrow$ & R$\uparrow$ \\ \midrule
CAS(ME)\textsuperscript{2} & - & - & 2.58 & 4.45 & 0.96 & 2.72 & 3.74 & 0.93 & 10.90 & 16.31 & 0.30 \\ \midrule
Celeb-DF (v2) & - & - & 13.60 & 19.07 & 0.13 & 23.24 & 28.55 & -0.08 & 16.23 & 21.27 & 0.11 \\ \midrule
\multirow{11}{*}{\begin{tabular}[c]{@{}c@{}}CAS(ME)\textsuperscript{2}\\ + Celeb-DF (v2)\end{tabular}} & - & - & 5.97 & 10.83 & 0.72 & 9.38 & 15.89 & 0.52 & 13.61 & 19.23 & 0.19 \\ \cline{2-12} 
 & Random & - & 3.28 & 4.91 & 0.96 & 7.69 & 15.20 & 0.41 & 10.93 & 17.93 & 0.42 \\ \cline{2-12} 
 & \multirow{4}{*}{Top/Bottom $K$} & COVER & 2.83 & 4.36 & 0.96 & 4.45 & 9.39 & 0.79 & 4.46 & 11.17 & 0.66 \\
 &  & SNR & 2.13 & 2.78 & 0.98 & 3.33 & 7.30 & 0.87 & 3.32 & 7.70 & 0.82 \\
 &  & IPR & 2.87 & 4.93 & 0.95 & 4.85 & 9.20 & 0.82 & 5.86 & 13.17 & 0.54 \\
 &  & \textbf{rPPG-VQA} & 2.19 & 2.55 & \underline{0.99} & 2.81 & 5.64 & 0.92 & 2.49 & 5.63 & 0.90 \\ \cline{2-12} 
 & \multirow{4}{*}{WRS} & COVER & 2.67 & 4.03 & 0.96 & 3.10 & 6.15 & 0.91 & 2.56 & 5.69 & 0.90 \\
 &  & SNR & 2.29 & 2.79 & \underline{0.99} & \underline{2.37} & \textbf{3.61} & \textbf{0.97} & 3.90 & 8.57 & 0.78 \\
 &  & IPR & 2.88 & 4.90 & 0.96 & 6.52 & 12.13 & 0.67 & 5.72 & 12.67 & 0.59 \\
 &  & \textbf{rPPG-VQA} & \underline{1.97} & \textbf{2.13} & \textbf{1.00} & 2.50 & 4.18 & \underline{0.96} & \underline{2.34} & \underline{5.16} & \underline{0.92} \\ \cline{2-12} 
 & TAS & \textbf{rPPG-VQA} & \textbf{1.86} & \underline{2.15} & \textbf{1.00} & \textbf{2.19} & \underline{3.99} & \underline{0.96} & \textbf{1.82} & \textbf{2.13} & \textbf{0.99} \\ \bottomrule
\end{tabular}
}
\end{table*}

\begin{table}
\caption{\textbf{Impact of augmenting training data with ``in-the-wild'' videos.} The Contrast-Phys method is used for model training.}
\label{tab:wild_data_effect}
\centering
\scalebox{0.7}{
\begin{tabular}{ccccc}
\toprule
\textbf{Training Set} & \textbf{Test Set} & \textbf{MAE$\downarrow$} & \textbf{RMSE$\downarrow$} & \textbf{R$\uparrow$} \\ \midrule
PURE & \multirow{2}{*}{PURE} & 1.00 & 1.40 & 0.99 \\
CAS(ME)\textsuperscript{2} + Celeb-DF (v2) + PURE &  & \textbf{0.15} & \textbf{0.18} & \textbf{1.00} \\ \midrule
PURE & \multirow{2}{*}{BUAA} & 2.43 & 3.25 & 0.97 \\
CAS(ME)\textsuperscript{2} + Celeb-DF (v2) + PURE &  & \textbf{1.76} & \textbf{1.98} & \textbf{1.00} \\ \midrule
BUAA & \multirow{2}{*}{BUAA} & 1.93 & 2.14 & \textbf{0.99} \\
CAS(ME)\textsuperscript{2} + Celeb-DF (v2) + BUAA &  & \textbf{1.12} & \textbf{1.59} & \textbf{0.99} \\ \midrule
BUAA & \multirow{2}{*}{PURE} & 2.74 & 5.53 & 0.93 \\
CAS(ME)\textsuperscript{2} + Celeb-DF (v2) + BUAA &  & \textbf{1.90} & \textbf{2.31} & \textbf{0.98} \\ \bottomrule
\end{tabular}
}
\end{table}

\subsection{Main Results}

\Cref{tab:results_pure} and \Cref{tab:results_buaa} show a comprehensive comparison of quality criteria and sampling strategies on the PURE and BUAA datasets, respectively, leading to several key observations.

\noindent \textbf{Superiority of rPPG-VQA Quality Criterion.} Across all sampling methods and downstream models, our proposed rPPG-VQA quality criterion consistently yields the best performance. When paired with our TAS strategy, models trained on data selected by rPPG-VQA achieve substantially lower MAE and RMSE and higher R values. This highlights the advantage of our holistic assessment, which evaluates both signal-level SNR and video-level scene noise. In contrast, criteria like SNR alone are effective but incomplete, while IPR often underperforms because it cannot distinguish between signal and noise harmonics.

\noindent \textbf{Effectiveness of TAS.} Our TAS strategy consistently delivers the best results when combined with our rPPG-VQA quality criterion, clearly outperforming other sampling methods like Top/Bottom $K$ and WRS. The Top/Bottom $K$ method struggles with skewed quality distributions, as its rigid selection rule forces the inclusion of low-quality data while underutilizing high-quality videos. In contrast, TAS employs a more nuanced two-stage process: it first filters out the lowest-quality candidates and then uses a duration-aware probabilistic mechanism to favor higher-quality videos. This promotes both quality and diversity, providing a distinct advantage over deterministic filtering and standard WRS.

\noindent \textbf{Value of ``in-the-wild'' Data.} Our experiments demonstrate that augmenting high-quality lab data with curated ``in-the-wild'' videos yields significant benefits. Training on the CAS(ME)\textsuperscript{2} dataset alone establishes a respectable performance baseline. However, simply adding more data is not the solution. For instance, training solely on the full Celeb-DF(v2) dataset fails to produce a viable model, and naively combining it with CAS(ME)\textsuperscript{2} severely degrades performance. In contrast, augmenting the lab data with a subset of Celeb-DF (v2) curated by our method leads to substantial improvements in accuracy. This result confirms an important finding: while a high-quality data foundation is essential, the advantages of diverse, real-world data can only be unlocked through an elaborate curation strategy that filters out detrimental noise. 

To further isolate this effect, we conducted a supplementary experiment (\Cref{tab:wild_data_effect}). We first established a baseline by training and testing a model on a single clean dataset (\eg, PURE). We then augmented the training set with our curated ``in-the-wild'' corpus. The results are striking: augmenting the training data caused a dramatic drop in error rates across both intra-dataset and the more challenging cross-dataset evaluations. This demonstrates that the diversity introduced by ``in-the-wild'' videos acts as a powerful regularizer, forcing the model to learn fundamental physiological features rather than overfitting to the idiosyncrasies of a single dataset.

\subsection{Ablation Studies}

\begin{table}
\caption{\textbf{Ablation study on the components of the signal-level noise perception branch.}}
\label{ablation_signal_branch}
\centering
\scalebox{0.7}{
\begin{tabular}{ccccc}
\toprule
\textbf{\begin{tabular}[c]{@{}c@{}}Frequency Consistency\\ Weight\end{tabular}} & \textbf{\begin{tabular}[c]{@{}c@{}}Spectral Correlation\\ Weight\end{tabular}} & \textbf{MAE$\downarrow$} & \textbf{RMSE$\downarrow$} & \textbf{R$\uparrow$} \\ \midrule
\ding{55} & \ding{55} & 0.82 & 1.47 & \textbf{1.00} \\
\checkmark & \ding{55} & 0.52 & 0.79 & \textbf{1.00} \\
\ding{55} & \checkmark & 0.57 & 0.90 & \textbf{1.00} \\
\checkmark & \checkmark & \textbf{0.47} & \textbf{0.74} & \textbf{1.00} \\ \bottomrule
\end{tabular}
}
\end{table}

\begin{table}
\caption{\textbf{Ablation study on the components of the TAS strategy.}}
\label{ablation_tas}
\centering
\scalebox{0.8}{
\begin{tabular}{ccccc}
\toprule
\textbf{\begin{tabular}[c]{@{}c@{}}Ranking-based\\ Filtering\end{tabular}} & \textbf{\begin{tabular}[c]{@{}c@{}}Duration-aware\\ Sampling\end{tabular}} & \textbf{MAE$\downarrow$} & \textbf{RMSE$\downarrow$} & \textbf{R$\uparrow$} \\ \midrule
\ding{55} & \ding{55} & 0.71 & 1.60 & 0.99 \\
\checkmark & \ding{55} & 0.70 & 1.22 & \textbf{1.00} \\
\ding{55} & \checkmark & 0.49 & 0.82 & \textbf{1.00} \\
\checkmark & \checkmark & \textbf{0.47} & \textbf{0.74} & \textbf{1.00} \\ \bottomrule
\end{tabular}
}
\end{table}

\begin{table}
\caption{\textbf{Ablation study on the fusion weight $\alpha$.}}
\label{tab:ablation_alpha}
\centering
\scalebox{0.7}{
\begin{tabular}{cccc}
\toprule
$\alpha$ & \textbf{MAE$\downarrow$} & \textbf{RMSE$\downarrow$} & \textbf{R$\uparrow$} \\ \midrule
0.0 & 6.91 & 14.06 & 0.49 \\
0.2 & 5.59 & 12.11 & 0.63 \\
0.4 & 4.39 & 10.81 & 0.70 \\
0.6 & 1.52 & 6.54 & 0.89 \\
0.8 & \textbf{0.47} & \textbf{0.74} & \textbf{1.00} \\
1.0 & 0.78 & 1.15 & \textbf{1.00} \\ \bottomrule
\end{tabular}
}
\end{table}

We conducted several ablation studies on the PURE dataset using the Contrast-Phys model to validate our framework's design choices.

\noindent \textbf{Component Analysis of Signal-Level Noise Perception Branch.} \Cref{ablation_signal_branch} dissects the contributions of the two weighting components in our signal-level noise perception branch. The baseline, a simple average of SNR scores, performs poorly. Incorporating either the frequency consistency or spectral correlation weight individually yields significant improvements. The complete framework, which combines both, achieves the best performance, confirming that our fusion of frequency and spectral consensus produces a more robust SNR estimate than simple averaging.

\noindent \textbf{Component Analysis of TAS.} In \Cref{ablation_tas}, we evaluate the components of our TAS strategy. The baseline (WRS without filtering or duration-awareness) is clearly outperformed by our full method. While adding ranking-based filtering provides a marginal benefit, incorporating duration-aware sampling offers a more substantial improvement. The full TAS strategy, which integrates both, delivers the optimal result. This demonstrates that the synergy between preliminary filtering and duration-aware probabilistic sampling is key to balancing data quality and diversity.

\noindent \textbf{Impact of Fusion Weight $\alpha$.} \Cref{tab:ablation_alpha} shows the model's sensitivity to the fusion weight $\alpha $, which balances the signal-level (${q_{sig}}$) and scene-level (${q_{sce}}$) scores. Relying on ${q_{sce}}$ alone ($\alpha  = 0.0$) is ineffective, confirming that scene information is not a substitute for signal analysis. Performance steadily improves as weight shifts to ${q_{sig}}$, with the optimum at $\alpha  = 0.8$. This synergistic configuration significantly outperforms using ${q_{sig}}$ alone ($\alpha  = 1.0$), validating our thesis that signal analysis and scene reasoning are complementary. The scene score is crucial for identifying physiologically invalid videos that purely signal-based metrics would otherwise miss.




\section{Conclusion}
\label{sec:Conclusion}

We address the pivotal but overlooked challenge of VQA for unsupervised rPPG. We propose rPPG-VQA, a novel framework shifting the focus from perceptual quality to physiological signal integrity. By fusing signal analysis with scene reasoning from an MLLM, our method robustly assesses a video’s suitability for rPPG. Our results demonstrate that leveraging rPPG-VQA to curate ``in-the-wild'' data significantly improves unsupervised methods. Models trained on this curated data achieve substantially higher accuracy than those trained solely on standard benchmarks. This work establishes that rPPG-specific VQA is not just beneficial but indispensable for unlocking the potential of large-scale, unlabeled video corpora. We hope our results will inspire the design of future video quality metrics to advance unsupervised rPPG.
\section*{Acknowledgements}
This work was supported by the National Natural Science Foundation of China under Grant 62172381.
{
    \small
    \bibliographystyle{ieeenat_fullname}
    \bibliography{main}
}

\clearpage
\setcounter{page}{1}
\maketitlesupplementary

\section{RANSAC Algorithm}

The random sample consensus (RANSAC) algorithm is an iterative method for robustly estimating the parameters of a mathematical model from data containing outliers~\cite{fischler1981random}. Its primary strength is its ability to derive a reliable model even when a significant fraction of the dataset consists of erroneous measurements. In this work, we employ RANSAC to determine a single, robust consensus frequency $f_{i,cons}^{HR}$, from a set of $M$ individual estimates $\left\{ {f_{i,j}^{HR}} \right\}_{j = 1}^M$.

RANSAC operates by assuming the dataset contains both inliers (data that can be explained by the model) and outliers (data that do not fit the model). The algorithm iteratively generates model hypotheses from random data subsets and evaluates them based on the support they receive from the full dataset. We demonstrate the overall algorithm for RANSAC in \Cref{alg:ransac}.

\begin{algorithm}
\caption{RANSAC Algorithm}
\label{alg:ransac}
\begin{algorithmic}[1]
\Require 
    \State $\mathcal{F} = \left\{ {f_{i,1}^{HR},...,f_{i,M}^{HR}} \right\}$: HR frequency estimates
    \State $\epsilon$: inlier tolerance threshold
    \State $K$: number of iterations 
\Ensure The estimated consensus HR frequency $f_{i,cons}^{HR}$
\State $s \gets \min \left( {2,M} \right)$
\State $I^* \gets \varnothing$
\For{$k = 1$ to $K$}
    \State $J \subseteq \left\{ {1,...,M} \right\}$ where $\left| J \right| = s$
    \State $\hat f_i^{HR} \gets \frac{1}{s}\sum\limits_{j \in J} {f_{i,j}^{HR}} $
    \State ${I_k} \gets \left\{ {j \in \left\{ {1,...,M} \right\}|\left| {f_{i,j}^{HR} - \hat f_i^{HR}} \right| \le \epsilon } \right\}$
    \If{$\left| {{I_k}} \right| > \left| {{I^*}} \right|$}
        \State ${I^*} \gets {I_k}$
    \EndIf
\EndFor
\If{$I^* \neq \varnothing$}
    \State $f_{i,cons}^{HR} \gets \mathrm{median} \left( {\left\{ {f_{i,j}^{HR}|j \in {I^*}} \right\}} \right)$
\Else
    \State $f_{i,cons}^{HR} \gets \mathrm{median} \left( \mathcal{F} \right)$
\EndIf
\State \Return $f_{i,cons}^{HR}$
\end{algorithmic}
\end{algorithm}

\section{Scene-Level Noise Perception Prompt}
\label{sec:Scene-Level Noise Perception Prompt}

\begin{figure*}[t]
    \centering
    \includegraphics[width=0.95\linewidth]{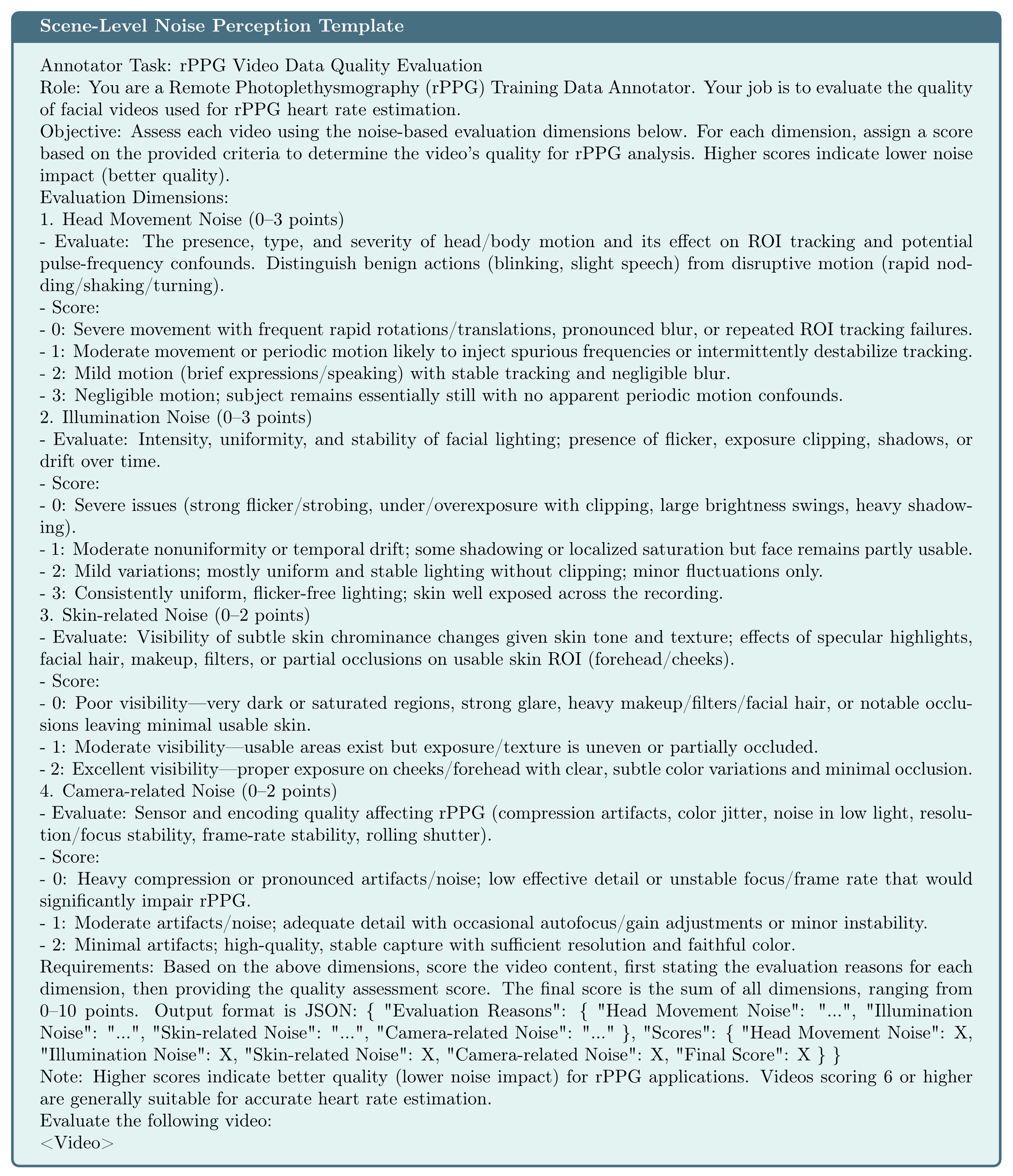}
    \caption{\textbf{Prompt for Qwen3-VL to assess scene-level quality.}}
    \label{fig:scene_prompt}
\end{figure*}

The prompt for Qwen3-VL~\cite{bai2025qwen3} to assess scene-level quality is given in \Cref{fig:scene_prompt}.

\section{WRS Algorithm}

Weighted random sampling (WRS) is a class of algorithms for drawing items from a collection, where each item's probability of being selected is proportional to an assigned weight~\cite{efraimidis2006weighted}. To construct the target training set ${{\mathcal{D}_{tgt}}}$ by resampling a source dataset ${\mathcal{D}_{src}}$, WRS effectively sample items with high-quality scores.

WRS first calculates the expected sampling count $r\left( {{v_i}} \right)$ for each data point ${v_i} \in {\mathcal{D}_{src}}$ using a softmax distribution over its quality score $Q\left( {{v_i}} \right)$, and scaled by the desired size of the target dataset $\left| {{\mathcal{D}_{tgt}}} \right|$:
\begin{equation}
    r\left( {{v_i}} \right) = \frac{{\exp \left( {Q\left( {{v_i}} \right)/\tau } \right)}}{{\sum\limits_{j \in {\mathcal{D}_{src}}} {\exp \left( {Q\left( {{v_j}} \right)/\tau } \right)} }}\left| {{\mathcal{D}_{tgt}}} \right|
\end{equation}
where the temperature parameter $\tau  > 0$, controls the sharpness of the sampling distribution.

The real-valued count $r\left( {{v_i}} \right)$ is then converted into an integer number of samples via stochastic rounding, ensuring the expected count for each item matches $r\left( {{v_i}} \right)$~\cite{xi2025samplemix}.

\section{Ablation Studies}

\subsection{Impact of Fusion Size $M$}

The results in \Cref{tab:ablation_m} demonstrate a clear correlation between the number of fused rPPG methods ($M$) and estimation accuracy. Relying on a single method ($M = 1$), results in the poorest performance. As we increase the fusion size from $M = 3$ to our chosen configuration of $M = 7$, we observe a consistent and significant reduction in both error metrics. This trend validates our core principle that a consensus-based fusion becomes more robust as it incorporates a larger set of diverse methods, effectively compensating for the idiosyncratic errors of individual rPPG estimators.

\begin{table}
\caption{\textbf{Ablation study on the fusion size $M$.}}
\label{tab:ablation_m}
\centering
\scalebox{0.95}{
\begin{tabular}{cccc}
\toprule
\textbf{$M$} & \textbf{MAE$\downarrow$} & \textbf{RMSE$\downarrow$} & \textbf{R$\uparrow$} \\ \midrule
1 & 0.91 & 1.30 & 0.99 \\
3 & 0.78 & 1.17 & 0.99 \\
5 & 0.57 & 1.12 & \textbf{1.00} \\
7 & \textbf{0.47} & \textbf{0.74} & \textbf{1.00} \\ \bottomrule
\end{tabular}
}
\end{table}

\subsection{MLLM Generalization and Stability}

We assessed generalization and stability by testing on Gemini 3.0 Pro and Kimi K2.5, using GPT-5.2 to generate perturbed prompt variations. \Cref{tab:mllm_generalization_stability} confirms our framework's robustness, showing minimal fluctuation in MAE, RMSE, and R across models and runs. Consequently, we retained Qwen3-VL as our primary model, as it provides the most favorable balance of latency and cost.

\begin{table}
\centering
\caption{\textbf{Generalization and stability analysis across different MLLMs.}}
\label{tab:mllm_generalization_stability}
\scalebox{0.9}{
\begin{tabular}{cccc}
\toprule
\textbf{MLLM} & \textbf{MAE$\downarrow$} & \textbf{RMSE$\downarrow$} & \textbf{R$\uparrow$} \\ \midrule
Qwen3-VL & 0.46 $\pm$ 0.02 & 0.77 $\pm$ 0.06 & 1.00 $\pm$ 0.00 \\
Gemini 3.0 Pro & 0.52 $\pm$ 0.05 & 0.73 $\pm$ 0.08 & 1.00 $\pm$ 0.00 \\
Kimi K2.5 & 0.55 $\pm$ 0.09 & 0.86 $\pm$ 0.11 & 1.00 $\pm$ 0.00 \\ \bottomrule
\end{tabular}
}
\end{table}

\section{Visualization}

\begin{figure}[t]
    \centering
    \includegraphics[width=0.95\linewidth]{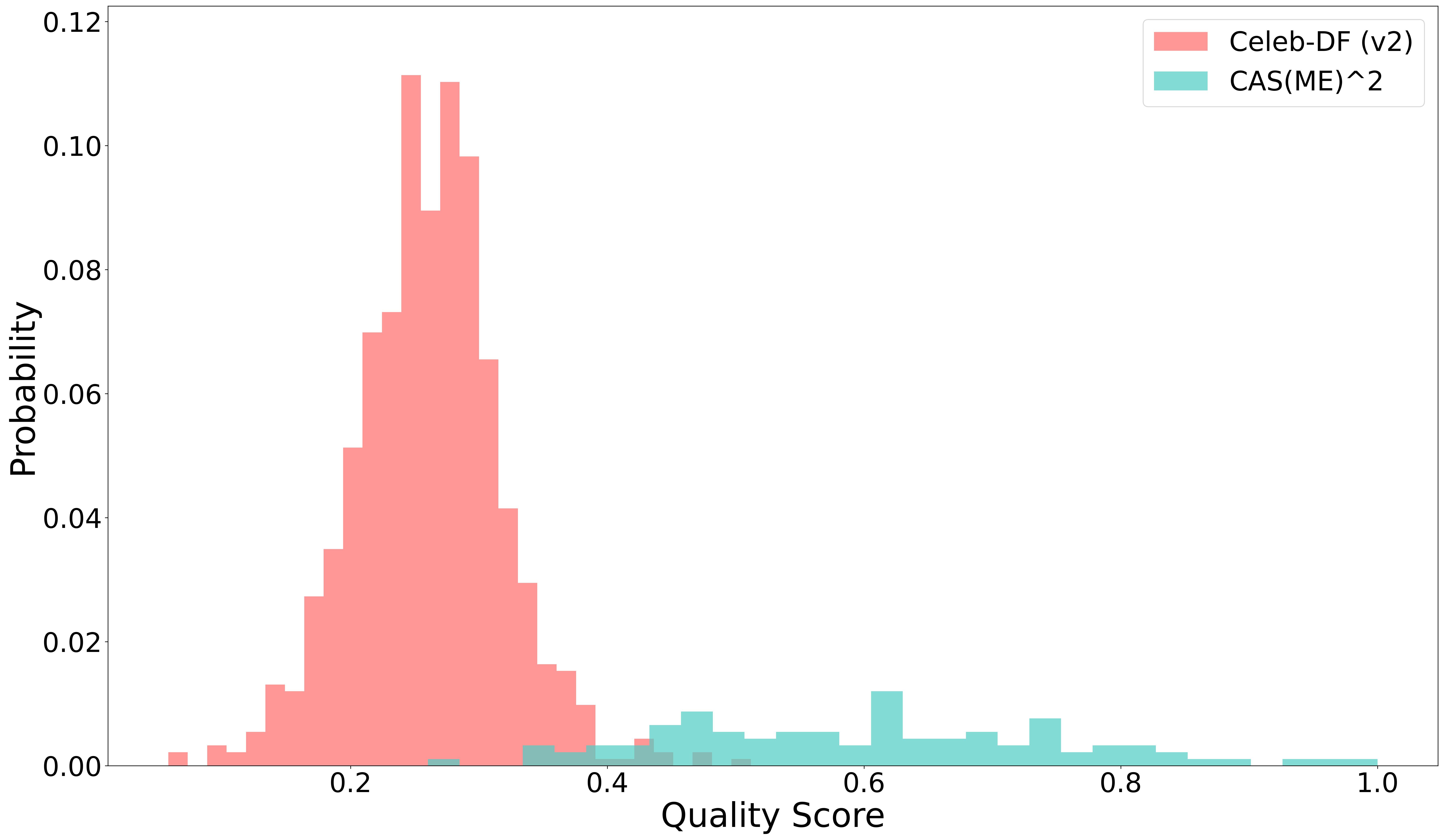}
    \caption{\textbf{Distribution of quality scores over the CAS(ME)\textsuperscript{2} and Celeb-DF (v2) datasets.}}
    \label{fig:quality_dist}
\end{figure}

\Cref{fig:quality_dist} illustrates a disparity in the quality score distributions between the CAS(ME)\textsuperscript{2}~\cite{qu2017cas} and Celeb-DF (v2)~\cite{li2020celeb} datasets. Scores for CAS(ME)\textsuperscript{2}, a controlled dataset, are concentrated in the high-quality range (0.4-0.8), reflecting its consistent signal fidelity. In contrast, scores for the ``in-the-wild'' Celeb-DF (v2) are predominantly clustered in the lower 0.1-0.4 range. While the latter offers valuable scenic diversity, it is plagued by poor signal quality. This quantitative analysis validates our core premise: a fundamental trade-off exists between data quality and diversity, necessitating a robust curation strategy to effectively leverage unvetted datasets.

\section{Failure Cases and Mitigation Mechanisms}

\subsection{Signal-level Branch Failure}

\Cref{fig:failure_cases}(a) illustrates a video from the ``in-the-wild'' MEVIEW dataset~\cite{husak2017spotting}. Existing estimators (GREEN, ICA, LGI, OMIT) produced inflated SNR values ranging from 16.85 to 26.69, leading to an erroneous consensus SNR of 20.72. This error likely stems from misinterpreting flickering background figures as physiological pulses. However, the scene-level branch effectively mitigated this by detecting the periodic visual noise, assigning a low quality score (6.00/10.00) to flag the sample.

\subsection{Scene-level Branch Failure}

\begin{figure}[t]
    \centering
    \includegraphics[width=0.95\linewidth]{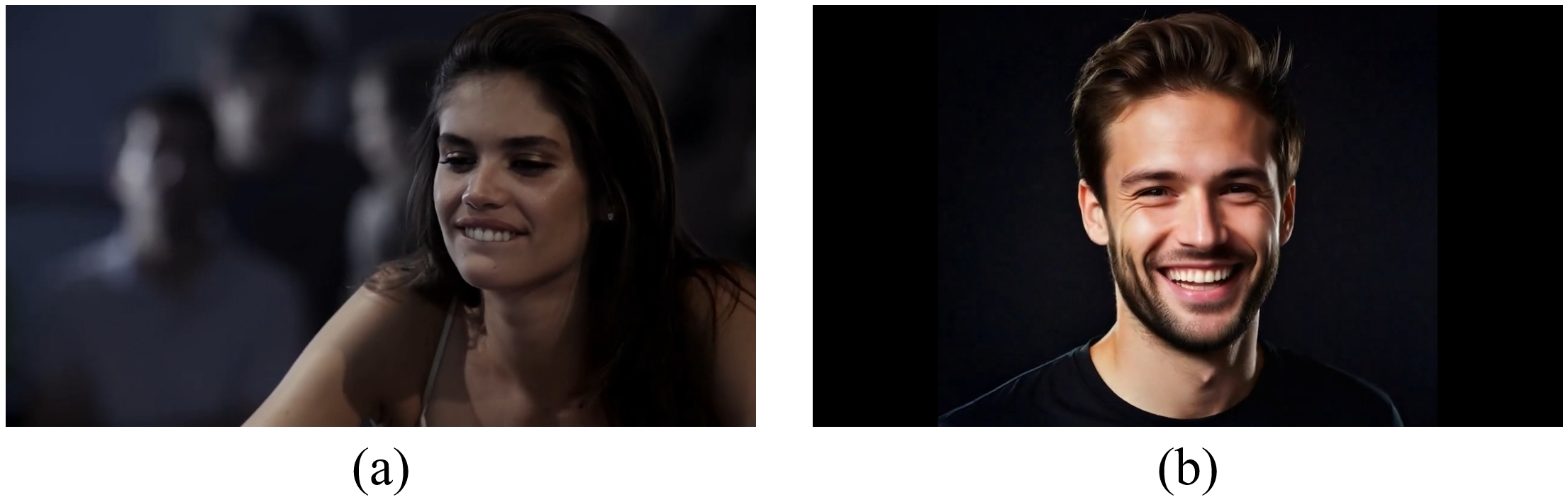}
    \caption{\textbf{Representative failure cases.} (a) Environmental noise interference; (b) AI-synthesized synthetic faces.}
    \label{fig:failure_cases}
\end{figure}

\Cref{fig:failure_cases}(b) depicts an AI-synthesized video void of rPPG signals. Here, the scene-level branch incorrectly assigned a high quality score (9.00/10.00) due to the video's visual stability. The signal-level branch compensated for this misclassification, yielding a consensus SNR of -2.67 and correctly identifying the absence of physiological signals.

\end{document}